\documentclass[lettersize,journal]{IEEEtran}
\usepackage{graphicx} % Required for inserting images
\usepackage{comment}
\usepackage{biblatex}
\usepackage[dvipsnames]{xcolor}
\usepackage{color, colortbl}

\usepackage{amsmath,amssymb,amsfonts}%
\usepackage{placeins}
\usepackage{multirow}
\usepackage{multicol}
\usepackage{booktabs}
\usepackage{fancyhdr}

\renewcommand{\thefootnote}{\fnsymbol{footnote}}
\title{DNN-based 3D Cloud Retrieval for Variable Solar Illumination and Multiview Spaceborne Imaging }
\author{Tamar Klein\textsuperscript{*}, Tom Aizenberg\textsuperscript{*}, Roi Ronen}
\begin{document}

\maketitle
\def\thefootnote{*}\footnotetext{These authors contributed equally to this work}

% Deep neural networks have become a popular solution to complex, non linear problems due to their effectiveness, speed, and relative accuracy. One such problem that benefits from the use of DNNs, and is further explored in this work, is the volumetric reconstruction of a three-dimensional heterogeneous scattering media, with unknown extinction coefficients per voxel. In particular, this work uses a DNN for the computational tomography (CT) of clouds, a method of volumetric recovery that utilizes multiple two-dimensional images of a scene from multiple perspectives to reconstruct the cloud's 3D properties.
\begin{abstract}
% ####
Climate studies often rely on remotely sensed images to retrieve two-dimensional maps of cloud properties. To advance volumetric analysis, we focus on recovering the three-dimensional (3D) heterogeneous extinction coefficient field of shallow clouds using multiview remote sensing data. Climate research requires large-scale worldwide statistics.
To enable scalable data processing, previous deep neural networks (DNNs) can infer at spaceborne remote sensing downlink rates. However, prior methods are limited to a fixed solar illumination direction.
In this work, we introduce the first scalable DNN-based system for 3D cloud retrieval that accommodates varying camera poses and solar directions. By integrating multiview cloud intensity images with camera poses and solar direction data, we achieve greater flexibility in recovery. Training of the DNN is performed by a novel two-stage scheme to address the high number of degrees of freedom in this problem. Our approach shows substantial improvements over previous state-of-the-art, particularly in handling variations in the sun's zenith angle.

\end{abstract}

\section{Introduction}

Clouds play a key role in the climate system~\cite{trenberth_2009}.
They form multi-scale, complex dynamical systems, with a multitude of feedbacks. However, clouds are challenging to resolve or represent in climate models.
A finer understanding of clouds is needed. We focus on retrieving the three-dimensional (3D) heterogeneous extinction coefficient field of shallow clouds using multiview remotely sensed data. 
% These small-sized clouds are sensitive to mixing with their dry environment. 
% This poses a great challenge to the cloud physics community~\cite{de_rooy_2013,gerber2000structure}.
% due to complex turbulent mixing and a lack of dense 3D atmospheric measurements~\cite{de_rooy_2013,gerber2000structure}.

% Current methods in remote sensing~\cite{nakajima1990determination} assume
% % data acquired by a single satellite. Then, 
% a {\em plane parallel} atmosphere structure to retrieve two-dimensional cloud information. The plane parallel approximation assumes that clouds are horizontally homogeneous, and therefore the radiative transfer (RT) model is mainly applied vertically. 
% This simplified assumption of horizontal homogeneity enabled sensing and computation when computers were weak, and satellites were very expensive. 
% This assumption leads to significant biases when applied to shallow scattered clouds, where 
% % it is known that 
% RT and cloud heterogeneity vary significantly in 3D~\cite{marshak20053d}. 

\begin{figure}[t]
    \centering
    \includegraphics[width=.99\linewidth]{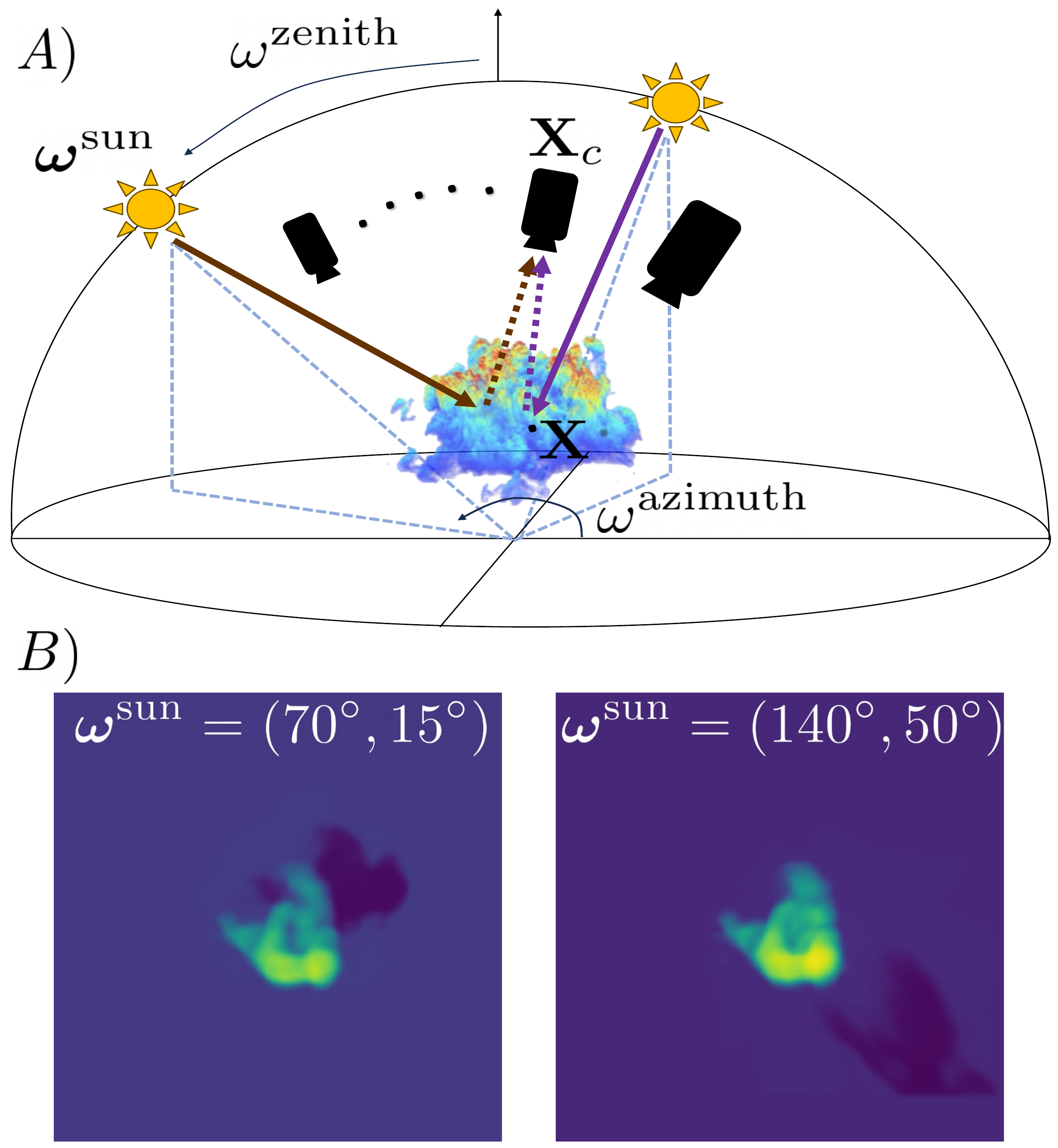}
    \caption{\emph{A}) A satellite formation in orbit abserves a cloud. The uncontrolled scene is illuminated by the sun in direction ${\boldsymbol{\omega}}^{\rm sun}=[\omega^{\rm zenith}, \omega^{\rm azimuth}]$. Radiation that reaches a point ${\bf X}$ in the atmospheric domain is multiply scattered until sensed by camera sensor at ${\bf X}_c$.  \emph{B}) Cloud single-channel intensity images are shown from the same satellite position.
    % with different solar radiation angles. 
    The images change significantly when illuminated from a different direction.
    }
    \vspace{-0.4cm}
    \label{fig:sat_formation}
\end{figure}

To sense clouds volumetrically in 3D, adequate observations are required as a prerequisite. This need led to the CloudCT space mission~\cite{schilling2019cloudct,tzabari2021cloudct}.
% , funded by the ERC. 
Cloud CT includes a coordinated {\em formation} of ten nano-satellites  {\em simultaneously} imaging a cloud field from multiple directions as illustrated in Fig.~\ref{fig:sat_formation}. 
In addition to observational improvements, advanced analysis is required.

% \begin{figure}[t]
%     \centering
%     \includegraphics[width=1\linewidth]{Figures/comparing images.jpg}
%     \caption{Example of images of the same cloud from the same camera perspective with different solar angles}
%     \label{fig:comparing_images}
% \end{figure}

Recent work~\cite{aides2020distributed,holodovsky2016situ,levis2015airborne,levis2017multiple,ronen20214d} showed the potential of cloud scattering tomography, which is a special form of computed tomography (CT). 
In contrast to X-ray CT,  cloud imaging is 
% a 
passive,
% imaging process
 relying on sun radiation as a source. Image readouts relate {\em nonlinearly} to volumetric cloud structure by 3D radiative transfer (RT): sunlight reaches the sensor by {\em scattering},
 % in the atmosphere,
 mainly from cloud droplets and atmospheric particles.
% forming the signal of raw image data.
To invert the 3D RT, Refs.~\cite{aides2020distributed,holodovsky2016situ,levis2015airborne,levis2017multiple,ronen20214d} use iterative-based optimization. However, these methods are slow and unscalable for extracting rich worldwide 3D cloud retrievals.

% To invert the 3D RT for volumetric recovery of clouds, Refs.~\cite{aides2020distributed,holodovsky2016situ,levis2015airborne,levis2017multiple} use iterative-based optimization for multi-view image data, given scene imaging setup. This includes the 3D camera poses and sun radiation direction.
% % . Additionally, these methods  prior knowledge of the 
% Optimization-based cloud CT methods
% % ,  while on the surface offers promising results, 
% have a major disadvantage: 3D RT is computationally slow to invert, as it is a nonlinear recursive operation. Therefore, optimization-based methods are unscalable for extracting rich worldwide 3D cloud retrievals.

To address recovery speed, deep neural network (DNN) methods were suggested~\cite{main_vipct,probct, nemf}.
% Due to the effectiveness, speed, and accuracy of DNNs, these networks have become a popular solution to complex, nonlinear problems.
% DNNs shift the computational burden to a {\em training stage} that can be executed at a time before imaging. Consequently, at inference, large data can be scalably analyzed at the downlink rates expected from spaceborne remote sensing. 
DNNs shift the computational load to a  {\em training stage}, making it feasible to analyse large data during inference, at rates comparable to downlink rates typical of spaceborne remote sensing.
% A DNN for cloud retrieval trains on a large
% , simulated, physics-based, 
% labeled dataset of cloud volumes and corresponding multi-view images.
% These simulated images must be rendered with variable imaging settings, as 
% Unlike X-ray CT, 
DNN-based cloud recovery requires training data of cloud volumes and corresponding multiview images. Cloud imaging is performed in an uncontrolled environment (see Fig.~\ref{fig:sat_formation}). Hence, data should include images with an arbitrary incoming sun direction and variable camera poses. The large number of degrees of freedom poses a challenge in acquiring such a large and diverse dataset. Furthermore, even if such a dataset exists, designing and training a DNN system for such complex data is not trivial.

% These challenges were addressed over a breakthrough line of work.
% First, a DNN-based system, termed 3DeepCT~\cite{3deepct}, was developed. 3DeepCT suggested using synthetic data. It was only trained for a fixed imaging geometry and therefore has no flexibility in the sun's direction or camera positions. 
% Its successor is VIP-CT (Variable Imaging Projection Cloud Tomography)~\cite{main_vipct}. VIP-CT has flexibility in camera position. Furthermore, the same trained system can process any atmospheric domain size, unlike 3DeepCT. However, despite the noted improvements, VIP-CT is mostly effective for a single illumination direction, the one on which it was trained. We show that VIP-CT performance significantly worsens when variability in solar angle is introduced.

In this work,
% we bridge this gap,
we address these challenges,
presenting Projection Integration for Variable OrienTation in Computed Tomography (PIVOT-CT): The first scalable system for 3D cloud retrieval with flexibility in both camera poses and sun direction. We accomplished this by (i) extending the VIP-CT~\cite{main_vipct} model by incorporating into its decoder the sun direction; and (ii) designing a novel two-stage training scheme. This scheme mitigates the challenge of the many degrees of freedom in our setup.

\section{Existing 3D Cloud Labeled Data}
\label{sec:background_deta}
Prior art~\cite{main_vipct,probct} for DNN-based cloud tomography requires labeled training data of volumetric clouds and their multiview images.
However, 
obtaining a large in-situ measured database
% of the extinction coefficients 
of volumetric cloud fields is infeasible. 
% Therefore, a large \emph{synthetic} 3D labeled cloud dataset is procured, and used in training, validating, and testing the networks.  
% Prior works~\cite{main_vipct,3deepct,nemf,probct}  used 
Therefore, a large \emph{simulated}  dataset of 3D cloud fields, termed BOMEX~\cite{bomex}, 
is procured.
% , and used in training, validating, and testing the networks.  
% The BOMEX dataset was generated by solving the physical thermodynamics equations for a turbulent atmosphere with real-world boundary conditions~\cite{khairoutdinov2003cloud}.
A cloud scene in this dataset is $1.6 \times 1.6$km wide and 1.2km thick, divided into 32 voxels along each axis. 
Per voxel, the cloud extinction coefficient ${\beta}^{\rm true}$, single scattering albedo, and phase function are known.
Training used $6000$ scenes. Testing used $566$ scenes. 

Per scene, the cloud field extinction coefficients ${\boldsymbol{\beta}}^{\rm true}$ and its multiview images are used as a ground-truth labeled pair.
Imaging geometry follows the CloudCT space mission:  $10$ satellites in a string-of-pearls formation, orbiting $500$km high.
The nearest neighboring satellite distance is $100$km. Imaging specifications are detailed in~\cite{main_vipct}.
% In real-world conditions, satellite position may be perturbed. Thus, VIP-CT introduced the { BOMEX${\rm _{perturbed}}$} dataset, which includes perturbations in the viewpoints geometry. There, each of the $10$ viewing satellites has a pose that is translated randomly.
% Let ${\cal U}[a,b]$ be the uniform distribution in the range $[a,b]$.
% Per camera in the formation, 
% each of the 3D spatial coordinates was drawn from ${\cal U}[-50{\rm km},50{\rm km}]$. 
% The randomly perturbed camera poses are known.
All synthetic images in the prior database
% for BOMEX and { BOMEX${\rm _{perturbed}}$} datasets 
were rendered using the SHDOM~\cite{levis2020git} RT solver, with a \emph{fixed} solar direction $\boldsymbol{\omega}_{\rm fixed}$, having a zenith angle of \mbox{${\omega}_{\rm fixed}^{\rm zenith}=30^\circ$} and an azimuth angle of ${\omega}_{\rm fixed}^{\rm azimuth}=135^\circ$.

\section{Methodology}

\subsection{Cloud Datasets with Uncontrolled Imaging}
% Varying Sun Direction}
\label{sec:datasets}

% In real-world conditions, satellite position may be randomly perturbed.
% Thus, VIP-CT introduced the { BOMEX${\rm _{perturbed}}$} dataset, which includes perturbations in the viewpoints geometry. There, each of the $10$ viewing satellites has a pose that is translated randomly.
% Let ${\cal U}[a,b]$ be the uniform distribution in the range $[a,b]$.
% Per camera in the formation, 
% each of the 3D spatial coordinates was drawn from ${\cal U}[-50{\rm km},50{\rm km}]$. 
% The randomly perturbed camera poses are known.

% Note from Eq.~(\ref{eq:decoder}) that the sun's direction is not an input to VIP-CT. 
% This was reasonable when VIP-CT was trained and tested on examples with a constant sun angle. 
% We show in Sec.~\ref{sec:results} that VIP-CT results degrade when tested on data with a varying sun direction, as is typical of a real-world scenario.
Remote sensing of clouds is done in an uncontrolled environment. Satellite positions may be randomly perturbed, and the solar direction \begin{equation}
    \boldsymbol{\omega}^{\rm sun}=[{\omega}^{\rm azimuth},{\omega}^{\rm zenith}]\;,
\end{equation} changes constantly relative to the clouds.
To generalize learned cloud tomography, training data should include varying imaging settings.
Hence, we introduce two new datasets: 
% BOMEX${\rm ^{sun}}$ and {BOMEX${\rm ^{sun} _{perturbed}}$}. 
% The datasets include training data to train our method (VIP-CT$^+$) and test examples for validation.
\\
% Similar to VIP-CT, we use a SHDOM generated dataset based on the BOMEX simulated 3D cloud fields. We generated three separate cloud datasets for our work.

% \subsubsection{BOMEX}
% The BOMEX dataset is generated identically to that used in VIP-CT training \ref{sec:background_deta}. 
% Solar azimuth and zenith angles are constant, and are $135^\circ$ and $45^\circ$ respectively.
% This dataset has 3075 training scenes, and 1052 test scenes.

\noindent {\textbf{BOMEX}${\rm ^{sun}}$}:
% \subsubsection{BOMEX-sun}
We randomly sample 1379 and 199 scenes from the BOMEX training and test sets, respectively.
Let ${\cal U}[a,b]$ be the uniform distribution in the range $[a,b]$.
For each training scene ${\boldsymbol{\beta}}^{\rm true}$, we drew ten times the solar direction by ${\omega}^{\rm azimuth}\sim{\cal U}[0,360^\circ]$ and ${\omega}^{\rm zenith}\sim{\cal U}[0,90^\circ]$.
Per sun direction,  ${\boldsymbol{\beta}}^{\rm true}$ is rendered using SHDOM, to obtain the multiview images under the sampled sun direction. 
% We follow the BOMEX dataset, described in Sec.~\ref{sec:datasets}, and used the same imaging geometry and image noise model. 
% For validation, each cloud test scene was rendered once by a random sun direction following this process.
\\

\noindent {\textbf{BOMEX}${\rm ^{sun} _{perturbed}}$}: This dataset follows the previous process to obtain cloud images with varying sun directions. In addition, in this dataset, the camera poses are perturbed: Per camera in the formation, 
each of the 3D spatial coordinates was randomly translated by  $\Delta x,\Delta y,\Delta z\sim{\cal U}[-50{\rm km},50{\rm km}]$. 
% as in BOMEX{${\rm _{perturbed}}$} (see Sec.~\ref{sec:datasets}).
This dataset has 845 training scenes and 566 test scenes. For both datasets, sun direction and camera poses are known. 

% The 
% For each sun direction, the s
% The BOMEX-sun dataset adds a variation in the solar azimuth and zenith angles. Here, the angles were distributed randomly in their respective ranges.
% This dataset has 1379 training scenes and 199 test scenes.

% \subsubsection{BOMEX-sun-perturbed}
% The BOMEX-sun-perturbed dataset adds an additional perturbation to the cameras.
% This dataset has 845 training scenes, and 567 test scenes.
%combined the two datasets we used.
%dataset structure for all 3. 10 cameras, -- sun angles, choose from those.

\begin{figure}[t]
    \centering
    \includegraphics[width=1\columnwidth]{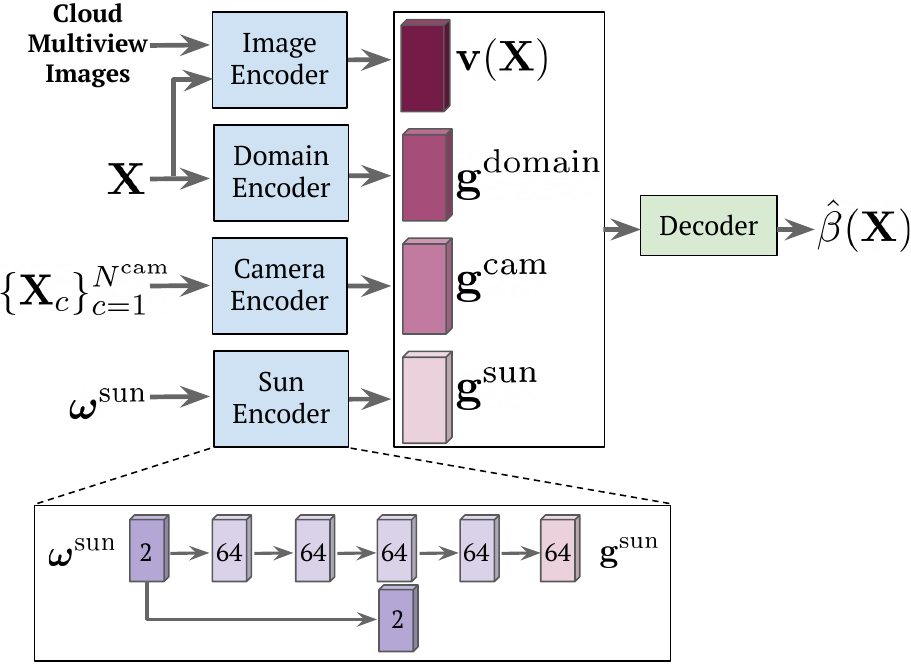}
    \caption{Architecture of PIVOT-CT. Feature extraction is performed on all images, resulting in a feature vector ${\bf v}({\bf X})$ for each atmospheric location ${\bf X}$, corresponding to the image pixels that are the geometric projections of ${\bf X}$.
    The atmospheric location ${\bf X}$ and camera positions $\{{\bf X}{c}\}_{c=1}^{N^{\rm cam}}$ are encoded into vectors ${\bf g}^{\rm domain}({\bf X})$ and $\{{\bf g}^{\rm cam}({\bf X}{c})\}_{c=1}^{N^{\rm cam}}$, respectively.   Additionally, the illumination direction $\boldsymbol{\omega}^{\rm sun}$ is provided to the system. A zoom-in of the Sun Encoder module is presented. The sun direction $\boldsymbol{\omega}^{\rm sun}\in {\mathbb R}^2$ is embedded to a vector ${\bf g}^{\rm sun}$ by five fully connected layers, each followed by a ReLU activation.
    % , facilitating improved performance on datasets with varying solar angles.
    These feature vectors are then input into a decoder, which predicts the extinction coefficient $\hat{\beta}({\bf X})$. 
% A skip connection is presented from the raw input vector to the third layer. 
}
    \label{fig:architecture}
\end{figure}
\subsection{PIVOT-CT}
\label{sec:vipct+}
%As previously mentioned, the VIP-CT network is flexible with regard to the inputs of camera pose and voxel location, and was trained using images with a constant solar radiance angle. 
The architecture of  PIVOT-CT is presented in Fig.~\ref{fig:architecture}.
PIVOT-CT  recovers the extinction coefficient per location of 3D heterogeneous clouds.
The inputs to PIVOT-CT are (i) $N^{\rm cam}=10$ multiview images of a cloud scene,
(ii) a 3D query location vector ${\bf X}$,  (iii) a vector of the camera array poses, denoted ${\{{\bf X}_{c}}\}_{c=1}^{N^{\rm cam}}$, and (iv) a vector of the sun direction $\boldsymbol{\omega}^{\rm sun}$.
% To encounter real-world scenarios of a varying sun direction, we incorporate 
% $\boldsymbol{\omega}^{\rm sun}$ to VIP-CT$^+$.

Following~\cite{main_vipct}, a convolutional neural network extracts input features of the images. The vector of features is denoted $\bf v$.
A query location ${\bf X}$ in the atmospheric domain can be projected to all camera planes,
using the known cameras' properties. 
The projected points on the camera planes are denoted $\{{\bf x}_{c}\}_{c=1}^{N^{\rm cam}}$.
Hence, per  ${\bf X}$, PIVOT-CT samples the features vector $\bf v$ at corresponding $\{{\bf x}_{c}\}_{c=1}^{N^{\rm cam}}$. This results in an image feature vector $\bf v(X)$.

Moreover, the vectors of the location ${\bf X}$,  camera positions ${\{{\bf X}_{c}}\}_{c=1}^{N^{\rm cam}}$, and sun direction $\boldsymbol{\omega}^{\rm sun}$ are processed by three individual fully connected-based DNNs that serve as geometry encoders.
This results in corresponding feature vectors ${\bf g}^{\rm domain}({\bf X})$, 
$\{
      {\bf g}^{\rm cam}({\bf X}_{c})
     \}_{c=1}^{N^{\rm cam}}$ and ${\bf g}^{\rm sun}(\boldsymbol{\omega})$, as seen in Fig.~\ref{fig:architecture}.
% The sun feature encoder is shown in Fig.~\ref{fig:sun_encoder}. 
     
These learned feature vectors encompass the spatial trends of the scene, influenced primarily by atmospheric physics and radiance.
By having these vectors as inputs, our system can handle the changes in viewing geometry and illumination direction.  
% \begin{figure}[t]
%     \centering
%     \includegraphics[width=1\linewidth]{Figures/sun_encoder.pdf}
%     \caption{Sun encoder module. The sun direction $\boldsymbol{\omega}^{\rm sun}\in {\mathbb R}^2$ is embedded by five fully connected layers, each followed by a ReLU activation.
% Following~\cite{mildenhall2020nerf}, a skip connection is presented from the raw input vector to the third layer. }
%     \label{fig:sun_encoder}
% \end{figure}
Overall, the encoded vectors are concatenated and fed to a decoder $D$ whose output is the estimated cloud extinction coefficient  $\hat \beta$ at  ${\bf X}$,
% of the geometry encoders are concatenated with the feature extractor output vector,
% to form the input to the main element of the system, the decoder. The decoder is an additional DNN whose output is the extinction coefficient of the corresponding 3D cloud, at the input voxel, and executes the function
% \begin{equation}
%   {\hat \beta}({\bf X}) 
%   =  D
%      \left[ {\bf v}({\bf X}),{\bf g}^{\rm domain}({\bf X}),
%      \{
%       {\bf g}^{\rm cam}({\bf X}_{c})
%      \}_{c=1}^{N^{\rm cam}}
%      \right]\;.
%   \label{eq:decoder}
% \end{equation}
\begin{equation}
  {\hat \beta}({\bf X}) 
  =  D
     \left[ {\bf v}({\bf X}),{\bf g}^{\rm domain}({\bf X}),
     \{
      {\bf g}^{\rm cam}({\bf X}_{c})
     \}_{c=1}^{N^{\rm cam}},
     {\bf g}^{\rm sun}(\boldsymbol{\omega})
     \right].
  \label{eq:decoder_+}
\end{equation}
% The architecture of the decoder is based on fully connected layers, similar to the VIP-CT decoder. 
% However, here the input dimension of the first layer of $D$ is larger by the size of the vector ${\bf g}^{\rm sun}$, 64 neurons.
% This increase in learnable weights is negligible in comparison to the total number of network parameters.

The 3D estimated cloud ${\hat {\boldsymbol{\beta}}}$ is obtained by querying in parallel all locations ${\bf X}$ which reside in the 3D cloud convex hull. The hull is estimated by space carving.

\begin{figure*}[t]
    \centering
    \includegraphics[width=1\linewidth]{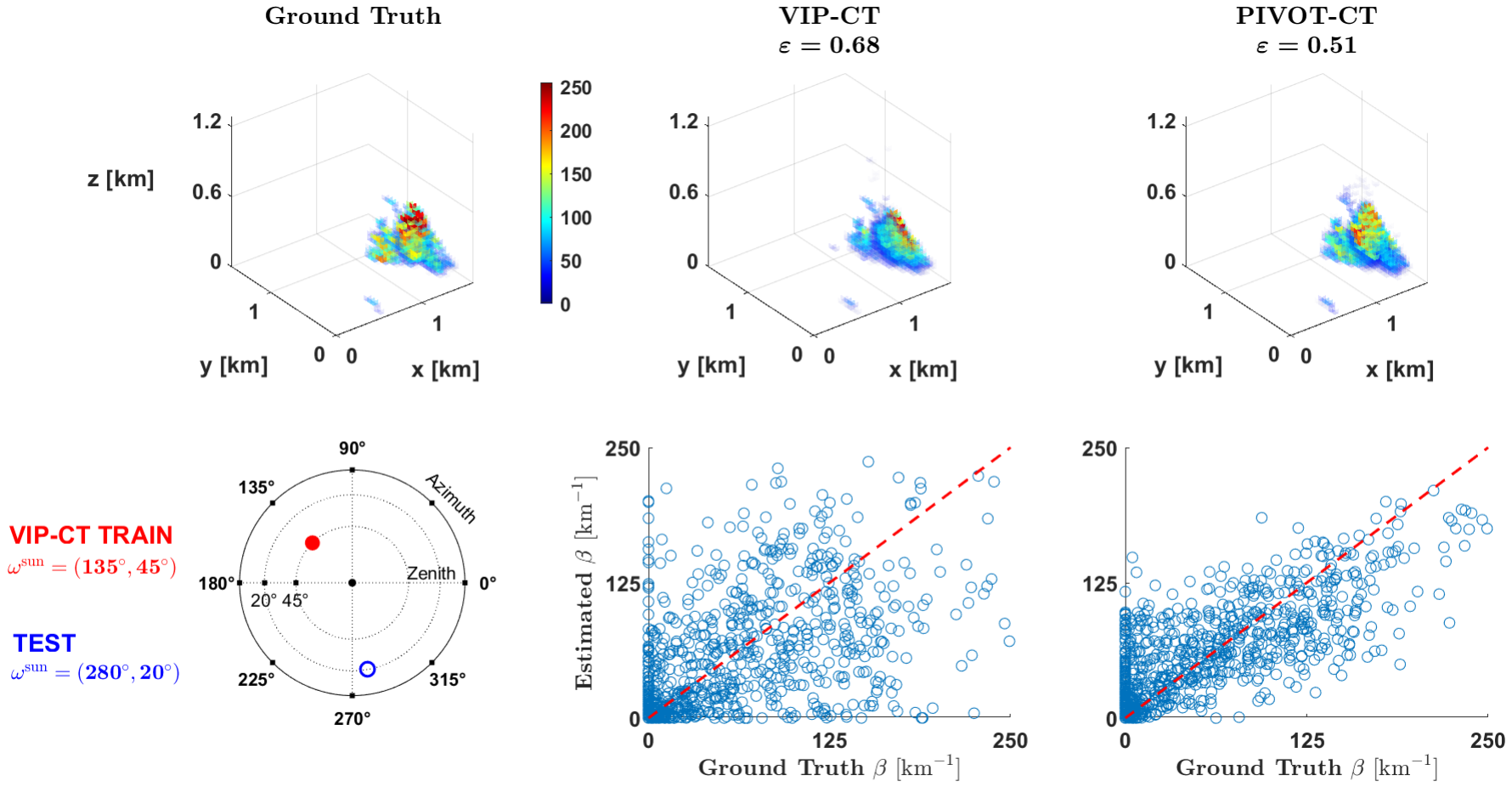}
    \caption{[Top] Visualizations of a test cloud's 3D extinction coefficient field and its recovery by VIP-CT and PIVOT-CT (ours). We show the relative error $\epsilon$ for each method.
  [Bottom] An illustration of the solar angle that VIP-CT was trained on and the true test scene solar angle. 
  Scatter plots of the estimated $\hat \beta$ compared to the true $\beta^{\rm true}$ across all voxels, the red dashed line represents optimal recovery.
    % row shows the ground truth $\beta$, and the estimations of both VIP-CT and VIP-CT$^+$. The bottom-left figure shows the difference between the solar angle VIP-CT was trained on and the solar angle of the data tested. It can be seen that VIP-CT$^+$ performed 17\% better than its predecessor VIP-CT. The two bottom right scatter plots show the estimated extinction coefficient $\beta$ vs $\beta_{gt}$
    }
    \label{fig:results}
\end{figure*}

\subsection{System Training}

Our DNN, PIVOT-CT, processes image data, imaging geometry, and the sun direction.
To handle the many degrees of freedom, we employ a two-stage training scheme.
In the first stage, PIVOT-CT parameters are randomly initialized. We freeze the sun encoder and set ${\bf g}^{\rm sun}\equiv0$. The network is trained until convergence on the BOMEX dataset.
% with a fixed image geometry and sun direction.

In the second training stage, we unfreeze the sun encoder ${\bf g}^{\rm sun}$ and train on data with varying sun direction, {{BOMEX}${\rm ^{sun} _{perturbed}}$} and {{BOMEX}${\rm ^{sun}}$}.
Network parameters are initialized by the previous stage weights.
% Here, we unfreeze the sun encoder ${\bf g}^{\rm sun}$ and train it additionally.
% To make training efficient, 
Here, we freeze the image feature extractor parameters.
Empirically, we found this
% that image encoder freezing in this stage 
improves performance. We follow~\cite{levis2015airborne, main_vipct} and numerically quantify the  per scene recovery quality by the relative error metric
\begin{equation}
 \epsilon = \left({\| { {\boldsymbol \beta}^{\rm true}} - {\hat {\boldsymbol \beta}} \|_1 }\right)
 {\Big /} {\Big (}{\| { {\boldsymbol \beta}^{\rm true}}  \|_1}{\Big )}
  \;.
  \label{eq:erros}
\end{equation}
We show in Table~\ref{tab:results} that our two-stage training scheme allows PIVOT-CT to better adapt to the complexity in {{BOMEX}${\rm ^{sun} _{perturbed}}$} and {{BOMEX}${\rm ^{sun}}$} datasets.

\subsubsection*{Implementation Details}

% The dataset used in training was generated by SHDOM, described in Sec \ref{sec:background_deta}. Each data point has ten 2D images of the same cloud, from the varying camera positions, and a random sun illumination direction. The sun azimuth and zenith angles have a uniform distribution in their respective ranges. 
Input sun angles $\omega^{\rm azimuth},\omega^{\rm zenith}$ are rounded to the nearest integer and normalized in the range $[0,1]$. 
% We found this normalization prevents model overfitting and makes training more efficient. 
% Additionally, we add slight noise to the data, to improve generalization. The camera positions are plus or minus 50 meters in 3D space.
Following~\cite{main_vipct}, each input image is normalized by the mean and standard deviation of the training data. 
We train to optimize the relative L2 norm as in~\cite{main_vipct}.
% the loss function 
% \begin{equation}
%  {\rm Loss}({ {\boldsymbol \beta}^{\rm true}},{\hat {\boldsymbol \beta}}) =\frac {\| { {\boldsymbol \beta}^{\rm true}} - {\hat {\boldsymbol \beta}} \|^2_2 }
%  {\| { {\boldsymbol \beta}^{\rm true}}  \|^2_2}
%   \;.
%   \label{eq:loss_mse}
% \end{equation}
System training stages endured $\approx 4e6$ and $0.25e6$ iterations, respectively. 
% The second stage ran additional $\approx 0.25e6$ iterations.  
Both stages use an Adam optimizer, a learning rate of $1e-5$, and a weight decay of $1e-5$. An iteration uses $200$ randomly sampled query voxels. 
The system runs on an NVIDIA GeForce RTX 3090 GPU.
% The test results presented in Table \ref{tab:results} are the mean and STD of $\epsilon$ over the data points of the test sets in the respective datasets \ref{sec:datasets}.

%processor: 12th Gen Intel Core i9-12900Kx24
%OS: 64 bit Ubuntu 22.04.4 LTS
%add mask?
% \FloatBarrier

\section{Results}

% \subsection{Results}
\label{sec:results}
% We present the test results of the VIP-CT$^+$ system.
Table~\ref{tab:results} compares PIVOT-CT with VIP-CT using the $\epsilon$ measure.
% We tested VIP-CT$^+$ on all three datasets in Sec \ref{sec:datasets}.
% The overall test results are summarized in Table ~\ref{tab:results}. 
% Due to the more general nature of VIP-CT$^+$, when tested on the BOMEX dataset, 
VIP-CT specializes in a fixed sun direction, thus it overpasses our method by $\approx 11\%$ on the fixed-sun BOMEX dataset.
% performed 
% $\approx 11\%$ better than VIP-CT$^+$. 
However, in a real-world scenario, where the sun's pose is random during imaging, PIVOT-CT outperforms VIP-CT on both BOMEX$^{\rm sun}$, BOMEX$^{\rm sun}_{\rm perturbed}$ datasets by $\approx 15\%$.
% on the remaining two more generalized datasets, the two-stage trained VIP-CT$^+$ outperformed VIP-CT by $\approx 11-15\%$. 
Also, PIVOT-CT fails when initialized randomly and directly trained on data with varying sun directions. This shows the benefit of the two-stage training process.
% produces worse results. 
Fig.~\ref{fig:results} shows a qualitative example from the BOMEX$^{\rm sun}_{\rm perturbed}$ dataset.

\begin{comment}
\begin{table}[h]
    \centering
    \renewcommand{\arraystretch}{1.5}
    \begin{tabular}{||c | c | c | c||} 
    \hline
       & BOMEX & BOMEX$^{sun}$ & BOMEX$^{sun}_{perturbed}$ \\  
     \hline\hline
     VIP-CT  & 34.53\% $\pm$ 9.56\% & 58.94\% $\pm$27.57\% & 61.54\% $\pm$ 27.99\% \\ 
     \hline
      VIP-CT$^-$  & 69.9\% $\pm$ 11\%  & 62.31\% $\pm$ 14.4\% & 62.75\% $\pm$ 13.77\%\\
    \hline
      VIP-CT$^+$  & 45.43\% $\pm$ 7.83\%  &   43.01\% $\pm$12.01\% & 46.31\% $\pm$ 13.58\% \\
      \hline \hline
    \end{tabular}
    \caption{Test Results: The metric used here is $\epsilon$ as shown in Eq. \ref{eq:erros} }
    \label{tab:results}
\end{table}
\end{comment}

\begin{table}[t]
\caption{Comparison of VIP-CT and PIVOT-CT (ours) using the $\epsilon$ measure (Eq.~\ref{eq:erros}). 
Our method outperforms VIP-CT on datasets having a varying sun direction.
two-stage training is very helpful.
}
\centering
\begin{tabular}{@{}ccccc@{}}
\toprule
                                             &                                                              & \multicolumn{3}{c}{Datasets} \\ \cmidrule(l){3-5} 
Method                                       & \begin{tabular}[c]{@{}c@{}}Two-Stage\\ Strategy\end{tabular} & \textbf{BOMEX}   & \textbf{BOMEX}${\rm ^{sun}}$   & \textbf{BOMEX}${\rm ^{sun} _{perturbed}}$  \\ \midrule
\multicolumn{1}{c|}{VIP-CT}                  & X                                                            & \textbf{35} $\pm$ \textbf{10}       & 58 $\pm$28        & 62 $\pm$ 28      \\ \midrule
\multicolumn{1}{c|}{\multirow{2}{*}{PIVOT-CT }} & X                                                            & 70 $\pm$ 11       & 62 $\pm$ 14        & 63 $\pm$ 14       \\
\multicolumn{1}{c|}{}                        & \checkmark                                                            & 45 $\pm$ 8       & \textbf{43} $\pm$ \textbf{12}        & \textbf{46} $\pm$ \textbf{14}       \\ \bottomrule
\end{tabular}
\label{tab:results}
\end{table}

\begin{figure}[h]
    \centering
    \includegraphics[width=1\linewidth]{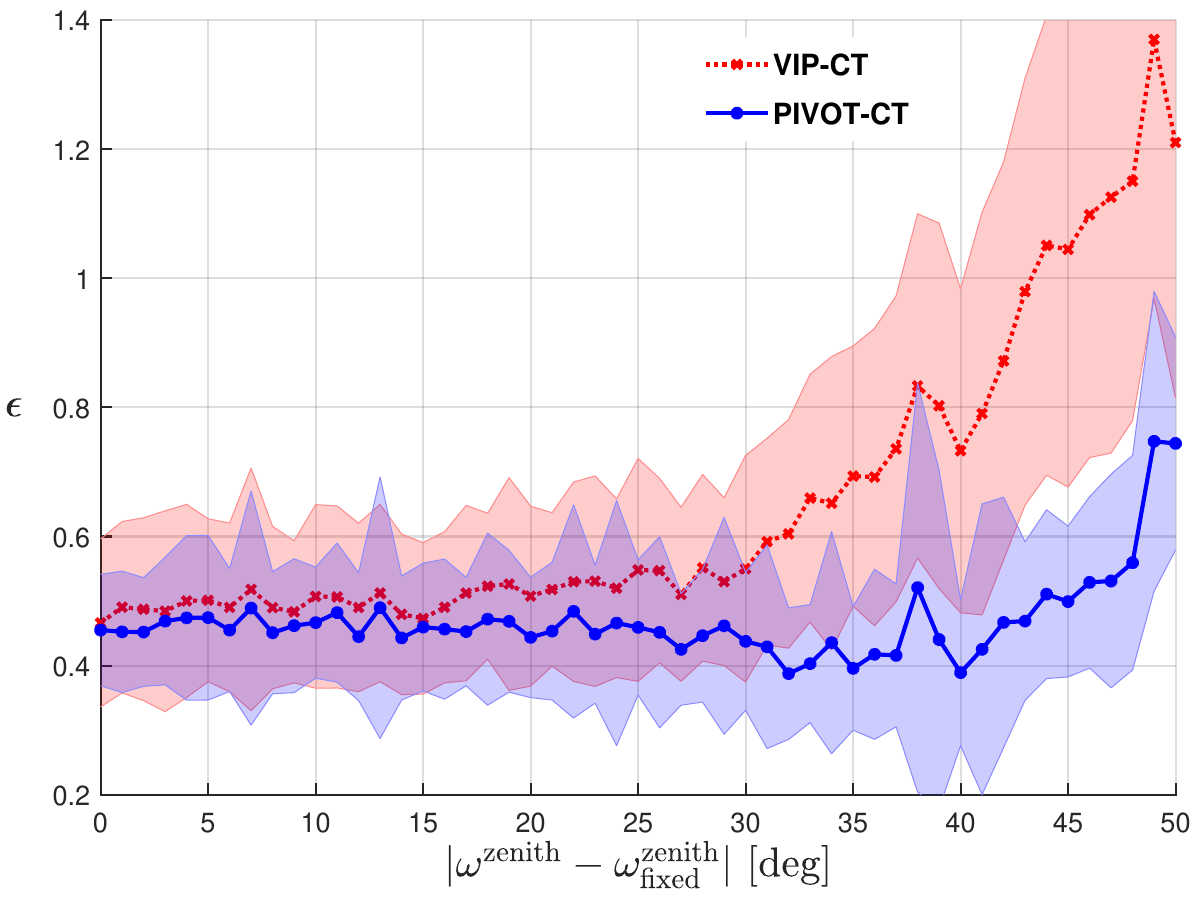}
    \caption{Comparison of the PIVOT-CT (ours) and VIP-CT systems. This figure shows the $\epsilon$ measure (Eq.~\ref{eq:erros}) as a function of the distance between the zenith angle of the test scenes and the zenith angle upon which VIP-CT was trained, $\omega^{\rm zenith}_{\rm fixed}$. The shaded area represents the standard deviation of $\epsilon$. The largest zenith angle difference of $50^\circ$ stands for the sun at the horizon.}
    \label{fig:shaded_error}
\end{figure}

Additionally, in Fig.~\ref{fig:shaded_error}, we compare the performance of  PIVOT-CT (ours) and VIP-CT  as a function of the sun zenith angle. 
% see from Figure \ref{fig:shaded_error} that the error of the VIP-CT system gets significantly larger the farther the test zenith angle is from $\omega^{\rm fixed} = 45^\circ$, 
% the VIP-CT training zenith angle. 
Our method surpasses prior art across all zenith angles, highlighting its robustness. 
% We stress that variation in the azimuth angle is less significant because solar illumination from a varying azimuth angle is akin to rotating the cloud around the vertical axis.

\section{Discussion}

We generalize learning-based cloud tomography to enable real-time volumetric retrieval of cloud fields using multiview imaging from realistic remote sensing. Our proposed method, PIVOT-CT, accommodates variable solar illumination angles and diverse camera array configurations.
PIVOT-CT demonstrates significant improvements over the state-of-the-art, particularly in handling variations in the sun's zenith angle.

As shown in Fig.~\ref{fig:shaded_error}, we observe a notable increase in tomographic error when the sun is near the horizon ($90^\circ$ from nadir). At this position, sunlight is attenuated by atmospheric particles, leading to dimmer cloud images and a reduced signal-to-noise ratio~\cite{HorizontalRadiative}. Additionally, when the sun is directly overhead, all views capture lower-order scattered light, whereas, in oblique sun position, more views capture higher-order scattered light. This results in decreased information content and less accurate retrievals.

Potential extensions of this work include exploring different integrations of the solar illumination direction, the incorporation of multimodal data, such as polarized imaging channels, and retrieval of a parameter vector at each location (e.g.,  droplet size distribution and density).

% The largest error of both VIP-CT and PIVOT-CT systems occurs when the zenith angle is farthest from \mbox{${\omega}_{\rm fixed}^{\rm zenith}$}, that is, the sun is at the horizon.
% We posit that the lingering trend is due to the pysical limitation of .
% %explain here why
% Moreover, while PIVOT-CT showed substantial improvement over VIP-CT in handling variations in the sun's zenith angle, its performance on real-world data could be further enhanced by incorporating additional physical factors. However, introducing these factors would increase the complexity of the learning task, making the training process more challenging and requiring more sophisticated training strategies.

\begin{comment}
% this was the old Discussion
The largest error of both VIP-CT and PIVOT-CT systems occurs when the zenith angle is farthest from \mbox{${\omega}_{\rm fixed}^{\rm zenith}$}, that is, the sun is at the horizon.
We posit that the lingering trend is a residual effect from VIP-CT, from which PIVOT-CT was initialized.
\end{comment}

\section{Acknowledgements}
We would like to express our sincere gratitude to Yoav Y. Schechner for his professional advice and Vadim Holodovsky for his help in procuring images.  We thank Daniel Yogodin, Ina Talmon, and Jochanan Erez for their technical support.
This project has received funding from the European Research Council (ERC) under the European Union’s Horizon 2020 research and innovation programme (CloudCT, grant agreement No. 810370), and the Israel Science Foundation (ISF grant 2514/23).

\section{Authors}
\noindent{Tamar Klein, Technion Israel Institute of Technology\\
Tom Aizenberg, Technion Israel Institute of Technology\\
Roi Ronen, Amazon Web Services\\}
\printbibliography
\end{document}